\pdfoutput=1
\documentclass[10pt,twocolumn,letterpaper]{article}
\usepackage[pagenumbers]{iccv} %

\usepackage{color}
\usepackage{times}
\usepackage{epsfig}
\usepackage{graphicx}
\usepackage{amsmath}
\usepackage{amssymb}
\usepackage{booktabs}
\usepackage{multirow}
\usepackage{arydshln}
\usepackage{pifont}
\usepackage{subcaption}
\usepackage{marvosym}
\usepackage{float}
\newcommand{\cmark}{\ding{51}}
\newcommand{\xmark}{\ding{55}}
\usepackage[table]{xcolor}
\newcommand{\thename}{OmniMamba}

\definecolor{iccvblue}{rgb}{0.21,0.49,0.74}
\usepackage[pagebackref,breaklinks,colorlinks,allcolors=iccvblue]{hyperref}

\title{OmniMamba: Efficient and Unified Multimodal Understanding and Generation via State Space Models}

\author{
Jialv Zou$^{1,\diamond}$ \quad
Bencheng Liao$^{2,1,\diamond}$ \quad
Qian Zhang$^{3}$ \quad 
Wenyu Liu$^{1}$ \quad 
Xinggang Wang$^{1,\textrm{\Letter}}$
\vspace{0.3em} 
\\
\fontsize{10.0pt}{9.84pt}
\textsuperscript{1} School of EIC, Huazhong University of Science \& Technology \\ 
\textsuperscript{2} Institute of Artificial Intelligence, Huazhong University of Science \& Technology \\ 
\textsuperscript{3} Horizon Robotics \\
}

\begin{document}

\twocolumn[
\maketitle
\vskip -0.3in
{
\vspace{-0.5cm}
\begin{center}
    \captionsetup{type=figure}
    \includegraphics[width=0.92\textwidth]{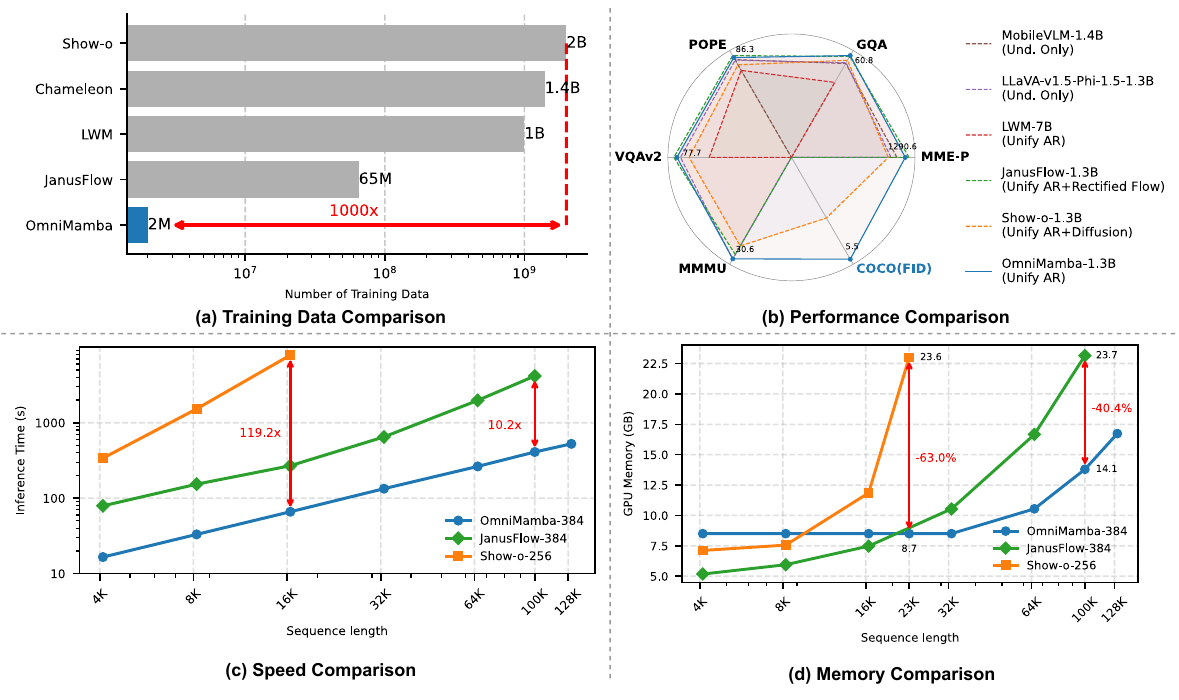}
    \vspace{-.1 in}
    \captionof{figure}{
        \textbf{Comprehensive comparison between \thename{} and other unified understanding and generation models.} 
        \textbf{(a)} Our \thename{} is trained on only 2M image-text pairs, which is 1000 times less than Show-o. 
        \textbf{(b)} With such limited data for training, our \thename{} significantly outperforms Show-o across a wide range of benchmarks and achieves competitive performance with JanusFlow. Black metrics are for the multimodal understanding benchmark, while the blue metric is for the visual generation task.
        \textbf{(c)-(d)} We compare the speed and memory of \thename{} with other unified models on the same single NVIDIA 4090 GPU. \thename{} demonstrates up to a 119.2$\times$ speedup and 63\% GPU memory reduction for long-sequence generation.
    }
    \label{fig:teaser}
\end{center} %
}]

\let\thefootnote\relax\footnotetext{$^\diamond$ Intern of Horizon Robotics.}
\let\thefootnote\relax\footnotetext{$^\textrm{\Letter}$ Corresponding author: \texttt{xgwang@hust.edu.cn}}

\begin{abstract}

Recent advancements in unified multimodal understanding and visual generation (or multimodal generation) models have been hindered by their quadratic computational complexity and dependence on large-scale training data. We present \thename{}, the first linear-architecture-based multimodal generation model that generates both text and images through a unified next-token prediction paradigm. The model fully leverages Mamba-2's high computational and memory efficiency, extending its capabilities from text generation to multimodal generation. To address the data inefficiency of existing unified models, we propose two key innovations: (1) decoupled vocabularies to guide modality-specific generation, and (2) task-specific LoRA for parameter-efficient adaptation. Furthermore, we introduce a decoupled two-stage training strategy to mitigate data imbalance between two tasks. Equipped with these techniques, \thename{} achieves competitive performance with JanusFlow while surpassing Show-o across benchmarks, despite being trained on merely 2M image-text pairs, which is 1,000 times fewer than Show-o. Notably, \thename{} stands out with outstanding inference efficiency, achieving up to a 119.2$\times$ speedup and 63\% GPU memory reduction for long-sequence generation compared to Transformer-based counterparts. Code and models are released at \url{https://github.com/hustvl/OmniMamba}

\end{abstract}

\section{Introduction}\label{sec:intro}
In recent years, Large Language Models (LLMs) ~\cite{dubey2024llama,touvron2023llama,touvron2023llama2,brown2020language,chowdhery2023palm} have achieved remarkable advancements, igniting significant research interest in extending their fundamental capabilities to the visual domain. Consequently, researchers have developed a series of Multimodal Large Language Models (MLLMs) for tasks such as multimodal understanding~\cite{liu2024visual,liu2024improved,zhu2023minigpt,zhou2024tinyllava} and visual generation~\cite{saharia2022photorealistic,koh2024generating}.

Recent studies have emerged that seek to integrate multimodal understanding with visual generation, aiming to develop unified systems capable of handling both tasks simultaneously. Such designs hold the potential to foster mutual enhancement between generation and understanding, offering a promising pathway toward truly unifying all modalities. 
Numerous studies have sought to preserve the text generation paradigm of LLMs while exploring the impact~\cite{xie2024show,wu2024janus,ma2024janusflow,wang2024emu3} of integrating diverse visual generation paradigms, such as diffusion models~\cite{ho2020denoising}, flow-based generative models~\cite{lipman2022flow,esser2024scaling}, and vector-quantized autoregressive models~\cite{sun2024autoregressive}.

Unfortunately, the significant domain gap between image and text presents a critical challenge for unified multimodal generative models: preserving generation capabilities without degrading understanding performance requires an extensive volume of image-text pairs for training, as illustrated in Fig.~\ref{fig:teaser}. This not only leads to poor training efficiency but also creates a substantial barrier to the broader development of such models, as only a small fraction of researchers possess the resources to undertake such computationally demanding studies. Moreover, most existing unified multimodal generative models rely on Transformer-based LLMs~\cite{vaswani2017attention}. However, their quadratic computational complexity results in slow inference speeds, rendering them less practical for real-time applications.
 
The challenges faced by existing unified multimodal generative models naturally lead us to ponder: \textbf{can a model be developed that achieves both training efficiency and inference efficiency?}

To address this, we introduce \thename{}, a novel unified multimodal generative model that requires only 2M image-text pairs for training. Built on the Mamba-2-1.3B~\cite{dao2024transformers} model as the foundational LLM with a unified next token prediction paradigm to generate all modalities, \thename{} leverages the linear computational complexity of state space models (SSMs) to achieve significantly faster inference speeds. Furthermore, to empower the Mamba-2 LLM—whose foundational capabilities are relatively weaker compared to the extensively studied Transformer models—to efficiently learn mixed-modality generation with limited training data, we propose novel model architectures and training strategies.

To enhance the model's capability in handling diverse tasks, we incorporate task-specific LoRA~\cite{hu2021lora}. Specifically, within each Mamba-2 layer's input linear projection, we introduce distinct LoRA modules for multimodal understanding and visual generation. During task execution, the features are modulated by both the linear projection and the corresponding task-specific LoRA, while the irrelevant LoRA components are deactivated. Furthermore, we propose the decoupled vocabularies to guide the model in generating the appropriate modality, which requires more data for the model to learn. On the data front, we further propose a novel two-stage decoupled training strategy to address the data imbalance between the two tasks, significantly improving training efficiency.

Trained on only 2M image-text pairs, our proposed \thename{} outperforms Show-o~\cite{xie2024show} on multiple multimodal understanding benchmarks and also matches the performance of JanusFlow~\cite{ma2024janusflow}, which was introduced by DeepSeek AI. Moreover, it achieves the best visual generation performance on the MS-COCO dataset~\cite{lin2014microsoft}. Notably, \thename{} demonstrates a 119.2$\times$ speedup at a sequence length of 16k and a 63\% GPU memory reduction at a sequence length of 23k, compared to Show-o. Furthermore, at a sequence length of 100k, it achieves a 10.2$\times$ speedup and 40.4\% memory savings compared to JanusFlow.
Our main contributions can be summarized as follows:
\begin{itemize}
    \item We introduce \thename{}, the first Mamba-based unified multimodal understanding and visual generation model to the best of our knowledge. By novelly adopting decoupled vocabularies and task-specific LoRA, \thename{} achieves effective training and inference.
    \item We propose a novel decoupled two-stage training strategy to address the issue of data imbalance between tasks. With this strategy and our model design, \thename{} achieves competitive performance using only 2M image-text pairs for training-up to 1,000 times fewer than previous SOTA models.
    \item Comprehensive experimental results show that \thename{} achieves competitive or even superior performance across a wide range of vision-language benchmarks and MS-COCO generation benchmark, with significantly improved inference efficiency, achieving up to a 119.2$\times$ speedup and 63\% GPU memory reduction for long-sequence generation on NVIDIA 4090 GPU.
\end{itemize}

\section{Related Work}\label{sec:related}
\begin{figure*}[htbp!]
    \centering
    \includegraphics[width=0.98\linewidth]{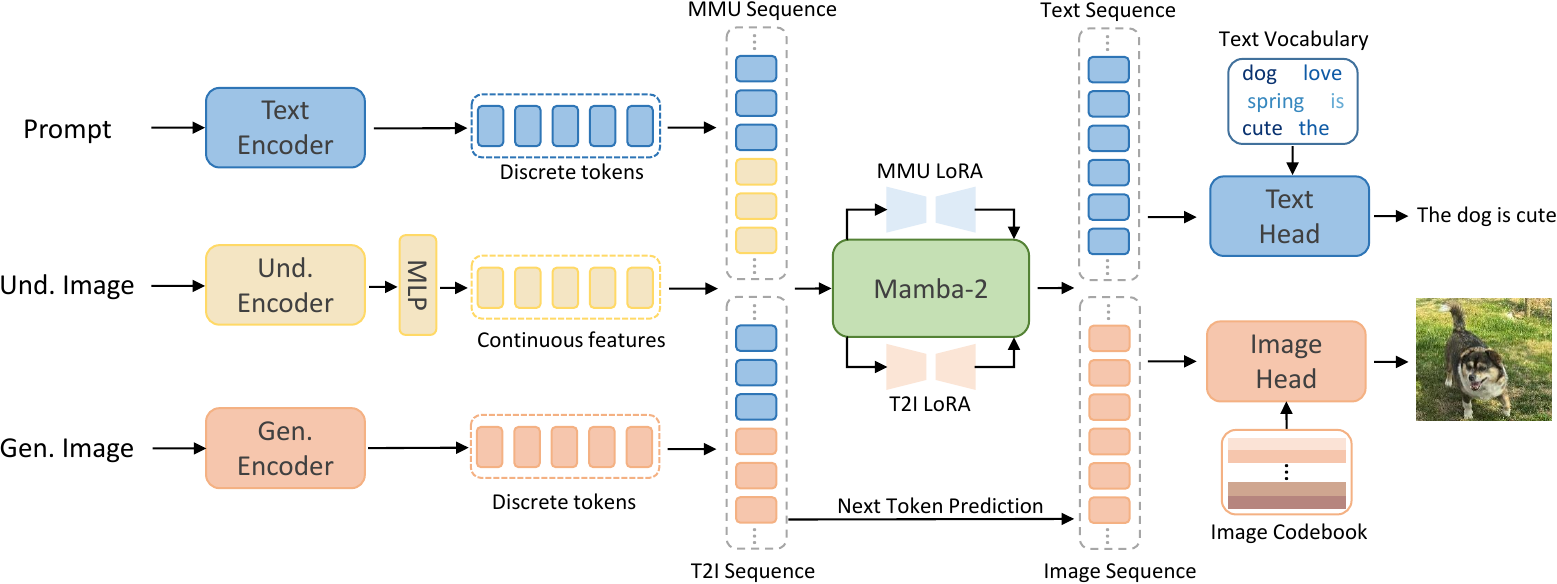}
    \caption{\textbf{Architecture of the proposed \thename{}} ``MMU'' refers to multimodal understanding, while ``T2I'' refers to text-to-image generation. \thename{} employs a next-token prediction paradigm for both multimodal understanding and visual generation tasks. To address the distinct requirements of each task—semantic information extraction for multimodal understanding and high-fidelity image compression for visual generation, we utilize separate encoders and heads. Furthermore, we purpose decoupled vocabularies to guide modality-specific generation and task-specific LoRA for parameter-efficient adaptation.}
    \label{fig:The overall architecture}
\end{figure*}

\paragraph{Multimodal Understanding} The remarkable advancements in LLMs have catalyzed the development of Large Vision-Language Models (LVLMs). Some representative works, such as the LLaVA series~\cite{zhu2024llava,liu2024visual}, BLIP series~\cite{li2022blip,li2023blip}, and MiniGPT-4~\cite{zhu2023minigpt}, have demonstrated strong multimodal understanding capabilities. These models align the features obtained from pretrained vision encoders with the feature space of LLMs through feature projectors, enabling pretrained LLMs to transfer their understanding and reasoning abilities to multimodal scenarios.

\paragraph{Visual Generation}
In recent years, diffusion models~\cite{ho2020denoising} have made remarkable progress, leading to models~\cite{rombach2022high,podell2023sdxl,dhariwal2021diffusion} with strong visual generation capabilities. Building on these advancements, flow-based generative models~\cite{lipman2022flow,esser2024scaling} have achieved superior results with fewer sampling steps. 
Additionally, some works~\cite{sun2024autoregressive,yu2024randomized} have successfully integrated autoregressive models into this domain, achieving notable performance.

\paragraph{Unified Understanding and Generation}
The remarkable advancements of LLMs in the fields of multimodal understanding and visual generation have naturally sparked researchers' interest in training a single LLM for both tasks. Early works~\cite{ge2024seed,dong2023dreamllm,ge2023planting,ge2023making} integrated pretrained diffusion modules as tools into LLMs, essentially forming a combination of two expert systems rather than utilizing a single LLM to perform both tasks. This approach results in a more complex model architecture and often leads to suboptimal outcomes. Show-o~\cite{xie2024show} integrates next-token prediction with discrete diffusion, enabling adaptive handling of mixed-modality inputs and outputs. Meanwhile, JanusFlow~\cite{ma2024janusflow} merges autoregressive models with rectified flow, a cutting-edge technique in visual generation. In contrast, Emu3~\cite{wang2024emu3} asserts that next-token prediction holds the greatest potential for achieving multi-modal generation, relying exclusively on this paradigm to manage both text and image generation tasks. Although these methods achieve outstanding performance, they are all based on Transformers, whose quadratic computational complexity presents a significant drawback, particularly when handling long-sequence generation tasks and high-resolution image generation. To address this challenge, we propose \thename{}, which employs a unified next token prediction paradigm to generate both text and image modalities, aiming to extend the linear computational complexity of the linear models to the field of multimodal generation.
\vspace{-0.3cm}
\paragraph{Linear Model} 
In recent years, a series of linear-complexity models have emerged as strong competitors to Transformers. Mamba~\cite{gu2023mamba}, a selective state space model, has garnered widespread attention for its competitive performance and faster inference speeds compared to Transformers. Building on this, Mamba-2~\cite{dao2024transformers}, an enhanced version of Mamba, achieves performance on par with Transformers while being 2-8 times faster. Similarly, GLA~\cite{yang2023gated} leverages gated linear attention to achieve linear complexity, maintaining competitive performance with Transformers.

The success of linear-complexity models in the field of natural language processing (NLP) has inspired its application in the visual domain, where it has been extensively studied in traditional image tasks~\cite{zhu2024vision,liu2024vmambavisualstatespace}, multimodal understanding~\cite{zhao2024cobra,huang2024ml,qiao2024vl,liao2025multimodal}, and visual generation~\cite{li2024scalable,hu2024zigma}. In this paper, we aim to design a unified multimodal understanding and visual generation model based on Mamba-2, which maintains competitive performance with Transformer-based unified models while offering significantly faster inference speeds.

\section{Method}\label{sec:method}
\subsection{Overall Architecture} 
Our ultimate goal is to design a unified multimodal understanding and visual generation model that achieves both training and inference efficiency using only 2M image-text pairs for training. We believe the key to realizing this goal can be summarized in one word: \textbf{decoupling}. To this end, we propose \thename{}, the architecture of which is illustrated in Fig.~\ref{fig:The overall architecture}.

Success necessitates standing on the shoulders of giants. We observe Emu3~\cite{wang2024emu3}, an autoregressive-based model which employs vast amounts of data and 8 billion model parameters. Despite these advantages, its final performance remains suboptimal, falling short of JanusFlow~\cite{ma2024janusflow}, a hybrid generative paradigm-based model with significantly less data and fewer parameters. We argue that this discrepancy stems not from the inherent superiority of the hybrid generative paradigm but from Emu3's tight coupling design, it uses the same vocabulary and encoder for all tasks and modalities. While this design aligns with the original intention of a unified model, it may lead to inefficient data utilization. In the following, we will introduce our model by focusing on the concept of decoupling.

\begin{figure}[t!]
    \centering
    \includegraphics[width=0.9\linewidth]{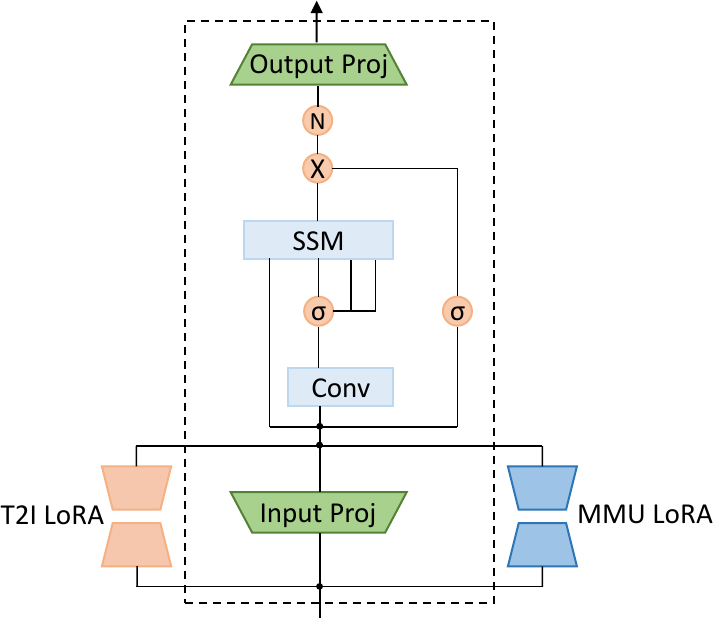}
    \caption{\textbf{The Mamba-2 block with task-specific LoRA}. It is worth noting that while the Mamba-2 Block in the Mamba-2 paper has two input projectors, the actual code implementation separates the feature dimensions from a single projector output. For simplicity, we depict only one input projector in our illustration. Our task-specific LoRA is applied to this entire input projector.}
    \label{fig:Mamba block}
    \vspace*{-0.5cm}
\end{figure}

\subsection{Decoupling Encoders for the Two Tasks} 
Previous works have explored using a single vision encoder for both tasks. For example, Show-o~\cite{xie2024show} employs MAGVIT-v2~\cite{yu2023language} to encode images into discrete tokens for both understanding and generation tasks. TransFusion~\cite{zhou2024transfusion} utilizes a shared U-Net or a linear encoder to map images into a continuous latent space for both tasks. Emu3 trains its vision encoder based on SBER-MoVQGAN5, enabling the encoding of video clips or images into discrete tokens.

However, JanusFlow~\cite{ma2024janusflow} has shown that such a unified encoder design is suboptimal. We believe this is primarily because multimodal understanding requires rich semantic representations for complex reasoning, whereas visual generation focuses on precisely encoding the spatial structure and texture of images. The inherent conflict between these two objectives suggests that a unified encoder design may not be the optimal choice. Therefore, \thename{} adopts a decoupled vision encoder design. 

Following prismatic VLMs~\cite{karamcheti2024prismatic}, we fuse DINOv2~\cite{oquab2023dinov2} and SigLIP~\cite{zhai2023sigmoid} as an encoder to extract continuous features for multimodal understanding. The key idea is that integrating visual representations from DINOv2, which capture low-level spatial properties, with the semantic features provided by SigLIP leads to further performance improvements. For visual generation, we use an image tokenizer trained with LlamaGen~\cite{sun2024autoregressive} to encode images into discrete representations. This tokenizer was pretrained on ImageNet~\cite{deng2009imagenet} and further fine-tuned on a combination of 50M LAION-COCO~\cite{laion2022} and 10M internal high aesthetic quality data.

\subsection{Decoupling Vocabularies for the Two Tasks} 
Unlike Emu3 and Show-o, which use a large unified vocabulary to represent both text and image modalities, to disentangle modality-specific semantics, we employ two separate vocabularies for each modality. This design explicitly separates the two modalities, providing additional modality-level prior knowledge. As a result, the model does not need to learn whether the output should be text or image, instead, it ensures the correct output modality by indexing the corresponding vocabulary. Our subsequent ablation experiments also confirm that \thename{}'s dual-vocabulary design is one of the key factors for efficient training.

\begin{figure*}[htbp!]
    \centering
    \includegraphics[width=0.9\linewidth]{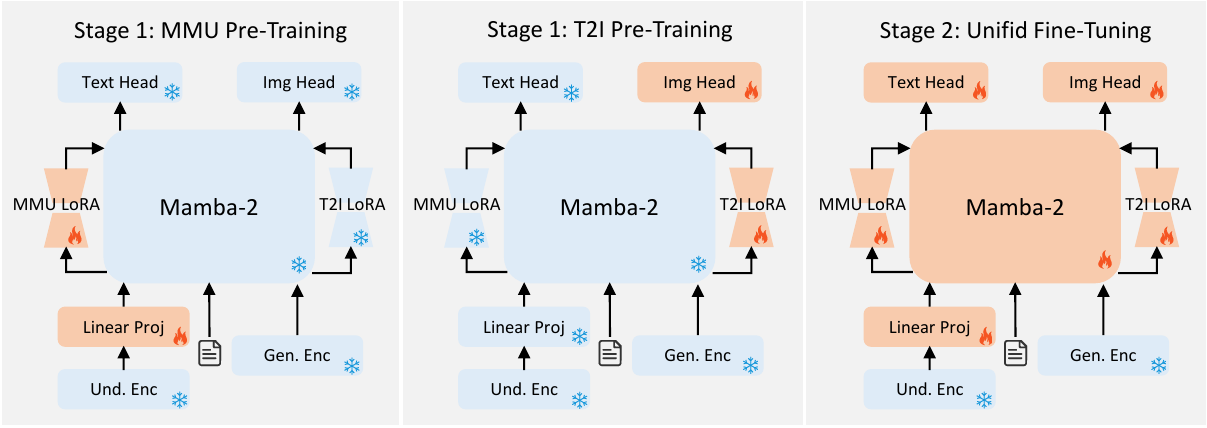}
    \caption{\textbf{Training strategy of \thename{}}. The trainable components are indicated by a flame symbol, while the frozen ones are represented by snowflakes. The dashed arrows indicate that this route is temporarily dropped and does not participate in model training.}
    \label{fig:The training strategy}
    \vspace*{-0.4cm}
\end{figure*}

\subsection{Task Specific LoRA}
To enhance the model's adaptability to specific tasks, we introduce task-specific adapters. We hypothesize that explicitly parameterizing the selection in SSMs based on task can enhance the data efficiency of multimodal training~\cite{dou2023loramoe}.  Specifically, to avoid introducing excessive parameters, we use LoRA~\cite{hu2021lora} as the adapter. In \thename{}, task-specific LoRA is applied only to the input projection of each Mamba-2 layer, as illustrated in Fig~\ref{fig:Mamba block}. When performing a specific task, the input linear projection and task-specific LoRA work together to effectively address the task. For instance, when the model performs a multimodal understanding (MMU) task, the MMU LoRA route is activated, while the text-to-image (T2I) LoRA route is dropped. Explicitly activating the corresponding adapter to assist in task execution helps improve data efficiency in training~\cite{dou2023loramoe,wang2023one}.

\subsection{Decoupled Training Strategy}
\label{sec:training_strategy}

We propose a decoupled two-stage training strategy to address data imbalance between understanding and generation tasks while improving training efficiency, as illustrated in Fig.~\ref{fig:The training strategy}. This approach consists of (1) a \textbf{Task-Specific Pre-Training} stage for module-specific initialization and modality alignment, and (2) a \textbf{Unified Fine-Tuning} stage for unified multi-task training.
\vspace{-0.2cm}
\paragraph{\textbf{Decoupling Rationale}}  
The first stage separates multimodal understanding (MMU) and text-to-image (T2I) generation tasks to prioritize modality alignment without data ratio constraints. Unlike joint pre-training methods (e.g., JanusFlow~\cite{ma2024janusflow} with a fixed 50:50 MMU-T2I data ratio), our approach trains task-specific modules independently, enabling flexible dataset scaling (665K MMU vs.\ 83K T2I samples). Only randomly initialized components---linear projection and MMU LoRA for understanding, T2I LoRA and image head for generation---are trained, while the core Mamba-2 model remains frozen. This eliminates competition between tasks during early learning and allows asymmetric data utilization.
\vspace{-0.2cm}
\paragraph{\textbf{Stage 1: Task-Specific Pre-Training}} It contains: \textbf{MMU Pre-Training}: Trains the linear projection and MMU LoRA to align visual-textual representations. The T2I LoRA path is disabled to isolate understanding-task learning. \textbf{T2I Pre-Training}: Optimizes the T2I LoRA and image decoder for visual synthesis. The MMU LoRA path is disabled to focus on generation capabilities.

\paragraph{\textbf{Stage 2: Unified Fine-Tuning}}  
Inspired by multi-task frameworks~\cite{karamcheti2024prismatic,zhu2024llava}, we freeze the visual encoder and train all other modules while preserving task-specific LoRA independence. During each forward pass:  (1) MMU and T2I computations use their respective LoRA branches; (2) Losses from both tasks are summed for a unified backward pass. This balances parameter sharing (via the frozen backbone) and task specialization (via isolated LoRA paths), enabling synergistic learning while mitigating interference between understanding and generation objectives.

\begin{table*}[h]
\centering
\setlength{\tabcolsep}{3.5pt}
\begin{tabular}{llccccccc}
\toprule
Type                       & Model              & LLM Params & Res. & POPE\(\uparrow\) & MME-P\(\uparrow\) & VQAv2$_{test}$\(\uparrow\)  & GQA\(\uparrow\)  & MMMU\(\uparrow\) \\ \midrule
\multirow{8}{*}{Und. Only} & LLaVA-Phi~\cite{zhu2024llava} & Phi-2-2.7B       & 336  & 85.0 & 1335.1 &71.4 & - & - \\
                           & LLaVA~\cite{liu2024visual} & Vicuna-7B       & 224  & 76.3 & 809.6 & - & - & - \\
                           & Emu3-Chat~\cite{wang2024emu3} & 8B from scratch       & 512  & 85.2 & - &75.1 & 60.3 & 31.6 \\
                           & LLaVA-v1.5~\cite{liu2024improved} & Vicuna-13B       & 448  & 86.3 & 1500.1 &81.8 & 64.7 & - \\
                           & InstructBLIP~\cite{dai2023instructblip} & Vicuna-13B       & 224  & 78.9 & 1212.8 & - & 49.5 & - \\
                           \cdashline{2-8}
                           & \cellcolor{blue!3}MobileVLM~\cite{chu2023mobilevlm} & \cellcolor{blue!3}MobileLLaMA-1.4B       & \cellcolor{blue!3}336  & \cellcolor{blue!3}84.5 & \cellcolor{blue!3}1196.2 & \cellcolor{blue!3}- & \cellcolor{blue!3}56.1 & \cellcolor{blue!3}- \\
                           & \cellcolor{blue!3}MobileVLM-V2~\cite{chu2024mobilevlm} & \cellcolor{blue!3}MobileLLaMA-1.4B       & \cellcolor{blue!3}336  & \cellcolor{blue!3}84.3 & \cellcolor{blue!3}1302.8 & \cellcolor{blue!3}- & \cellcolor{blue!3}59.3 & \cellcolor{blue!3}- \\
                           & \cellcolor{blue!3}LLaVA-v1.5-Phi-1.5~\cite{xie2024show} & \cellcolor{blue!3}Phi-1.5-1.3B       & \cellcolor{blue!3}336  & \cellcolor{blue!3}84.1 & \cellcolor{blue!3}1128.0 & \cellcolor{blue!3}75.3 & \cellcolor{blue!3}56.5 & \cellcolor{blue!3}30.7 \\
\midrule
\multirow{7}{*}{Unified}    & LWM~\cite{liu2024world}             & LLaMA2-7B       & 256  & 75.2 & - & 55.8 & 44.8 & - \\
                            & Chameleon~\cite{team2024chameleon}             & 7B from scratch       & 512  & - & - & - & - & 22.4 \\
                            & LaVIT~\cite{jin2309unified}             & 7B from scratch       & 256  & - & - & 66.0 & 46.8 & - \\
                            & Emu3~\cite{wang2024emu3}             & 8B from scratch       & 512  & 85.2 & 1243.8 & 75.1 & 60.3 & 31.6 \\
                            \cdashline{2-8}
                           & \cellcolor{blue!3}Janus~\cite{wu2024janus}          & \cellcolor{blue!3}DeepSeek-LLM-1.3B      & \cellcolor{blue!3}384  & \cellcolor{blue!3}87.0 & \cellcolor{blue!3}1338.0 & \cellcolor{blue!3}77.3 & \cellcolor{blue!3}59.1 & \cellcolor{blue!3}30.5 \\
                           & \cellcolor{blue!3}JanusFlow~\cite{ma2024janusflow}          & \cellcolor{blue!3}DeepSeek-LLM-1.3B      & \cellcolor{blue!3}384  & \cellcolor{blue!3}88.0 & \cellcolor{blue!3}1333.1 & \cellcolor{blue!3}79.8 & \cellcolor{blue!3}60.3 & \cellcolor{blue!3}29.3 \\
                           & \cellcolor{blue!3}Show-o~\cite{xie2024show}             & \cellcolor{blue!3}Phi-1.5-1.3B       & \cellcolor{blue!3}512  & \cellcolor{blue!3}80.0 & \cellcolor{blue!3}1097.2 & \cellcolor{blue!3}69.4 & \cellcolor{blue!3}58.0 & \cellcolor{blue!3}26.7 \\
                           & \cellcolor{blue!3}\textbf{\thename{}}          & \cellcolor{blue!3}Mamba-2-1.3B      & \cellcolor{blue!3}384  & \cellcolor{blue!3}86.3 & \cellcolor{blue!3}1290.6 & \cellcolor{blue!3}77.7 & \cellcolor{blue!3}60.8 & \cellcolor{blue!3}30.6 \\
\bottomrule
\end{tabular}
\caption{\textbf{Comparison with other methods on multimodal understanding benchmarks.} ``Und. only'' refers to models that only perform multimodal understanding task, while ``Unified'' refers to models that unify both multimodal understanding and visual generation tasks. Models with a similar number of parameters to ours are highlighted in light blue for emphasis.}
\label{tab:MMU}
\end{table*}

\subsection{Training Details}
\paragraph{\textbf{Data Formats}} Following Show-o~\cite{xie2024show}, we use special tokens to unify the data formats for both multimodal understanding and visual generation tasks. The multimodal understanding data is structured as:
\[
\small{\text{[MMU][SOI]\{image tokens\}[EOI][SOT]\{text tokens\}[EOT]}}.
\]
While the visual generation data is:
\[
\small{\text{[T2I][SOT]\{text tokens\}[EOT][SOI]\{image tokens\}[EOI]}}.
\]
Specifically, [MMU] and [T2I] is a pre-defined task token used to guide the model in performing the corresponding task. [SOT] and [EOT] are used to represent the beginning and end of text tokens, respectively. Similarly, [SOI] and [EOI] represent the beginning and end of image tokens.

\paragraph{\textbf{Training Objective}} Since \thename{} uses the auto-regressive paradigm to handle both multimodal understanding and visual generation tasks, we only need to use the standard cross-entropy loss for next-token prediction during training.

\section{Experiment}\label{sec:exp}

\subsection{Data}
To achieve the goal of data efficiency, we aim to train \thename{} using as few high-quality image-text pairs as possible.
\vspace{-0.2cm}
\paragraph{\textbf{Multimodal Understanding Data}} In the first pretrain stage, the training data consists of 676K image-text pairs, all of which are sourced from publicly available datasets. These includes 118K images from COCO~\cite{lin2014microsoft} and 558K images from LLaVA-1.5 pre-training data~\cite{liu2024improved}. 

In the second fine-tune stage, we also exclusively use publicly available datasets follow Cobra~\cite{zhao2024cobra}. 
\begin{enumerate}
    \item The mixed dataset used in LLaVA-1.5~\cite{liu2024improved} consists of 665K visual multi-turn conversations.
    \item LVIS-Instruct-4V~\cite{wang2023see}, which comprising 220K images accompanied by visually aligned and context-aware instructions generated by GPT-4V.
    \item LRV-Instruct~\cite{liu2023mitigating}, a large-scale visual instruction dataset of 400K samples, designed to mitigate hallucination issues across 16 vision-and-language tasks.
\end{enumerate}
\vspace{-0.2cm}
\paragraph{\textbf{Visual Generation Data}} To facilitate better reproducibility and further exploration by the community, we using only 83K images from the MS-COCO 2014 dataset~\cite{lin2014microsoft} for text-to-image generation training.

Overall, the training data for our unified multimodal generation model consists of fewer than 2 million image-text pairs. In contrast to previous works that rely on over 100M or even over 1B pairs, our approach is highly training efficient.

\subsection{Implementation Details}
Our core model is based on Mamba-2-1.3B, which consists of 48 layers of Mamba-2 blocks. In our primary experiments, the input image resolution for the multimodal understanding task is 384, while the image resolution for the visual generation task is 256.

\begin{table}[h]
\setlength{\tabcolsep}{1.5pt}
\centering

\begin{tabular}{llccc}
\toprule
Type                       & Model              & Params & Images & FID-30K\(\downarrow\)  \\ 
\midrule
\multirow{7}{*}{Gen. Only} & DALL·E~\cite{ramesh2021zero} & 12B       & 250M  & 27.5 \\
                            & GLIDE~\cite{nichol2021glide} & 5B       & 250M  & 12.24 \\
                            & DALL·E 2~\cite{ramesh2022hierarchical} & 6.5B       & 650M  & 10.39 \\
                            & SDv1.5~\cite{rombach2022high} & 0.9B       & 2000M  & 9.62 \\
                            & PixArt~\cite{chen2023pixart} & 0.6B       & 25M  & 7.32 \\
                            & Imagen~\cite{saharia2022photorealistic} & 7B       & 960M  & 7.27 \\
                            & Parti~\cite{yu2022scaling} & 20B       & 4.8B  & 7.23 \\
                            & Re-Imagen~\cite{chen2022re} & 2.5B       & 50M  & 6.88 \\
                            & U-ViT~\cite{bao2023all}     & 45M       & 83k(coco)  & 5.95 \\
\arrayrulecolor{black}\midrule
\multirow{7}{*}{Unified}    & CoDI~\cite{tang2024any}    & -       & 400M  & 22.26 \\
                            & SEED-X~\cite{ge2024seed}     & 17B       & -  & 14.99 \\
                            & LWM~\cite{liu2024world}    & 7B       & -  & 12.68 \\
                            & DreamLLM~\cite{dong2023dreamllm}     & 7B       & -  & 8.76 \\
                            & Show-o~\cite{xie2024show}     & 1.3B       & 35M  & 9.24 \\
                            & \textbf{\thename{}}    & 1.3B       & 83k(coco)  & \textbf{5.50} \\

\bottomrule
\end{tabular}
\caption{\textbf{Compare visual generation capability with other methods on MS-COCO validation dataset.} ``Gen. only'' refers to models that only perform visual generation task, while ``Unified'' refers to models that unify both multimodal understanding and visual generation tasks.}
\label{tab:T2I}
\end{table}

\begin{table}[h]
\centering
\begin{tabular}{lcc}
\toprule
Model     & $Gen_{avg}$ (Image/s) & $Total$ (s) \\
\midrule
Show-o~\cite{xie2024show}    & 0.81                    & 19.66      \\
JanusFlow~\cite{ma2024janusflow} & 1.02                    & 15.64      \\
\textbf{\thename{}}      & \textbf{5.68}                       & \textbf{2.81}        \\
\bottomrule
\end{tabular}
\caption{\textbf{Image Generation Speed in Visual Generation Task.} \thename{} achieves 7.0 $\times$ faster image generation speed compared to Show-o and 5.6 $\times$ faster compared to JanusFlow.}
\label{tab:speed}
\vspace{-0.3cm}
\end{table}

For multimodal understanding, we combine DINOv2~\cite{oquab2023dinov2} and SigLIP~\cite{zhai2023sigmoid} as the image encoder, while for visual generation, we use the VQVAE trained by LlamaGen~\cite{sun2024autoregressive} as the image encoder. We incorporate task-specific LoRA into the input projector of each Mamba-2 block and set the LoRA rank to 8, which results in only a $0.65\%$ increase in parameters. All training stages use the AdamW optimizer with $\beta_{1}$ set to 0.9 and $\beta_{2}$ set to 0.95. We adopt cosine annealing with warm-up as the learning rate schedule. Weight decay is set to 0, and gradient clipping is applied with a threshold of 1.0. Other detailed hyper-parameters are shown in the appendix. All of our training is conducted on NVIDIA A800 GPUs with BF16 precision.

\subsection{Quantitative Results}
\paragraph{\textbf{Multimodal Understanding}} We evaluate \thename{}'s multimodal understanding capabilities on a wide range of vision-language benchmarks, including POPE~\cite{li2023evaluating}, MME~\cite{fu2024mmecomprehensiveevaluationbenchmark}, GQA~\cite{hudson2019gqa}, MMMU~\cite{yue2024mmmu}. The results are shown in Tab~\ref{tab:MMU}. Compared to models with a similar number of parameters, \thename{} surpasses understanding-specific models such as LLaVA-v1.5-Phi-1.5~\cite{xie2024show}, MobileVLM~\cite{chu2023mobilevlm}, and MobileVLMv2~\cite{chu2024mobilevlm}. It also outperforms the unified understanding and generation model Show-o~\cite{xie2024show} and achieves competitive performance compare to the state-of-the-art unified model JanusFlow~\cite{ma2024janusflow}. Notably, while Show-o utilizes 2B image-text pairs and JanusFlow leverages over 65M image-text pairs, \thename{} achieves competitive performance by using only 2M image-text pairs for training.

\begin{figure*}[htbp!]
    \centering
    \includegraphics[width=0.9\linewidth]{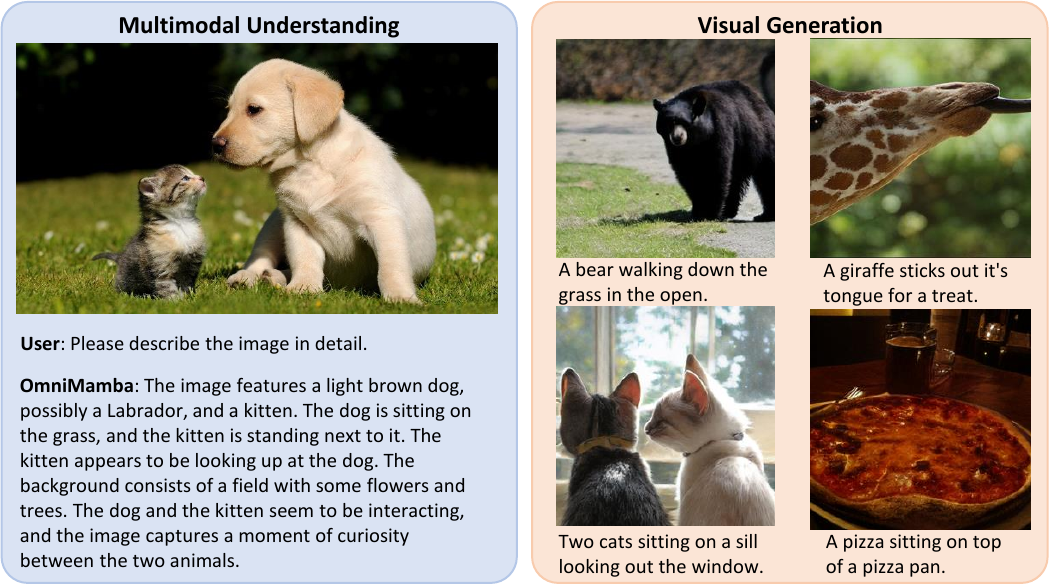}
    \caption{Qualitative results of \thename{} on multimodal understanding and visual generation.}
    \label{fig:qualitative}
\end{figure*}

\begin{table*}[h]
\centering
\begin{tabular}{ccccccc}
\toprule
\#Exp & Decoupling Vocabularies & Task Specific LoRA & POPE\(\uparrow\) & MME\(\uparrow\)  & GQA\(\uparrow\)  & FID-30K\(\downarrow\) \\
\midrule
1 & \xmark & \cmark & 80.8 & 1036 & 53.6 & 19.1    \\
2 & \cmark & \xmark & 81.2 & 1003 & 54.0 & 14.4    \\
3 & \cmark & \cmark  & \textbf{81.9} & \textbf{1100} & \textbf{55.3} & \textbf{10.3}   \\
\bottomrule
\end{tabular}
\caption{Ablation studies on decoupling Vocabularies and task specific LoRA in \thename{}}
\label{tab:abl}
\vspace{-0.2cm}
\end{table*}

\paragraph{\textbf{Visual Generation}} We evaluate \thename{} for text-to-image generation on the widely recognized MS-COCO~\cite{lin2014microsoft} benchmark dataset. To quantify image quality, we report the FID score~\cite{heusel2017gans}. Consistent with previous literature, we randomly select 30K prompts from the MS-COCO validation set and generate corresponding images to compute the FID score. The results are shown in Tab~\ref{tab:T2I}, where our model achieves the best visual generation performance on the MS-COCO validation dataset. Notably, models such as Show-o~\cite{xie2024show} and PixArt~\cite{yu2022scaling} are trained on external large-scale datasets and further fine-tuned on COCO-like datasets (e.g., OpenImages~\cite{kuznetsova2020open}) before evaluating zero-shot on the MS-COCO validation set. In contrast, to avoid introducing excessive additional data, both our model and U-ViT~\cite{bao2023all} are trained solely on the MS-COCO training set and evaluated on the MS-COCO validation set.

\subsection{Qualitative Results}
We present qualitative evaluations of our \thename{} for both multimodal understanding and visual generation tasks. Fig~\ref{fig:qualitative} showcases our model's capabilities in scene description and text-guided generation. Additional visualization results can be found in the appendix.

\subsection{Inference Speed and GPU Memory usage}
We compared the generation speed and GPU memory usage of OmniMamba with other Transformer-based models in both multimodal understanding and visual generation tasks. All the evaluations were done on the same single NVIDIA 4090 GPU with FP16 precision.

In multimodal understanding task, all models received the same example image. We used the same prompt, ``Please describe the image in detail.'' and removed the token generation limit to test their generation speed. The results are shown in the Fig~\ref{fig:teaser}. \thename{} demonstrates 119.2$\times$ speedup at a sequence length of 16k, and saves 63.0\% GPU memory at a sequence length of 23k compared to Show-o-256. Meanwhile, at a sequence length of 100k, \thename{} achieves a 10.2 $\times$ speedup and 40.4\% GPU memory savings compared to JanusFlow-384, which is accelerated by FlashAttention-2~\cite{dao2023flashattention}. Notably, Show-o-256 indicates that the input image resolution is 256. Due to the design of its omni-attention mechanism being incompatible with FlashAttention-2, it was not used during testing. Similarly, JanusFlow-384 represents an input image resolution of 384, with FlashAttention-2 applied during testing for acceleration.

In visual generation task, we used the same prompt, ``A Picture''. The models generate images with a batch size of 16 and a resolution of 256. As shown in Tab.~\ref{tab:speed}, our model achieves image generation speeds that are 7.0 $\times$ faster than Show-o and 5.6 $\times$ faster than JanusFlow.

\subsection{Ablation Studies}
We conducted a series of ablation studies to verify the effectiveness of each design in \thename{}. In this section, all ablation studies are conducted based on Mamba-2-370M, with the understanding visual encoder replaced by CLIP~\cite{radford2021learning}, an input resolution of 224, and a reduced number of training steps (kept consistent across all ablation experiments), while keeping all other settings unchanged.
\vspace{-0.2cm}
\paragraph{\textbf{Impact of Decoupling Vocabulary for the Two Tasks}} To guide the model in generating specific modalities from a structural design perspective, we employ modality-decoupled vocabularies in the default \thename{}. We conducted ablation studies on this design. As shown in Tab~\ref{tab:abl}, Exp1 and Exp3 utilize a modality-unified vocabulary and modality-decoupled vocabularies, respectively. The results demonstrate that the decoupled vocabularies enable more efficient model training and yield better performance. Notably, when using the modality-unified vocabulary for visual generation, the model occasionally produces text-related tokens, which requires additional post-processing to ensure correct visual generation. This further indicates that the model needs additional training to effectively learn modality-specific generation.
\vspace{-0.2cm}
\paragraph{\textbf{Impact of Task Specific Adapter}} We conducted ablation studies on the introduced task-specific adapter module, and the results are shown in Tab~\ref{tab:abl}. Exp2 and Exp3 represent the model without and with the task-specific adapter, respectively. The experiments demonstrate that the task-specific adapter helps the model efficiently learn both multimodal understanding and visual generation with a minimal amount of training image-text pair data (2M).

\section{Conclusion}\label{sec:conclusion}
We presented \thename{}, the first Mamba-2-based unified multimodal understanding and visual generation framework that achieves competitive performance with remarkable inference and training efficiency. By introducing three key innovations: decoupled vocabularies to disentangle modality-specific semantics, task-specific LoRA modules for parameter-efficient adaptation, and a two-stage decoupled training strategy to resolve data imbalance, \thename{} achieves comparable performance with JanusFlow and even surpasses Show-o using only 2M image-text pairs for training. Moreover, \thename{} exhibits outstanding inference efficiency, It achieves a 119.2$\times$ speedup with a sequence length of 16k and a 63\% reduction in GPU memory at a sequence length of 23k, compared to Show-o. With a sequence length of 100k, it delivers a 10.2$\times$ speedup and saves 40.4\% of GPU memory compared to JanusFlow. These results validate that our proposed \thename{} is both training and inference efficient, with the potential to enable more ordinary researchers to participate in the wave of unified model innovation. However, due to the limited scale of training data, our model's performance remains slightly below SOTA methods. Exploring the trade-off between training data volume and model performance will be a key focus of our future work.

{\small
\bibliographystyle{ieee_fullname}
\bibliography{OmniMamba}
}

\appendix
\section*{Appendix}
\section{Training Details}
The detailed training hyper-parameters are listed in Tab~\ref{tab:Hyper-parameters}. The first stage separates multimodal understanding (MMU) and text-to-image (T2I) generation tasks to prioritize modality alignment without data ratio constraints. To this end, we employ a larger learning rate during the first pre-training stage and a smaller learning rate in the second fine-tuning stage. The batch size ratio represents the proportion between multimodal understanding data and visual generation data. All training is conducted on NVIDIA A800 GPUs using BF16 precision.

\begin{table}[htbp!]
    \setlength{\tabcolsep}{2.5pt}
    \begin{tabular}{lccc}
    \toprule
                                        & Stage 1: MMU & Stage 1: T2I & Stage 2: Unify \\
    \midrule
    \multicolumn{1}{l|}{Learning Rate}  & 1e-3       & 8e-4       & 1e-4         \\
    \multicolumn{1}{l|}{Warm-up Steps}  & 100        & 1000       & 0            \\
    \multicolumn{1}{l|}{Training Steps} & 5k       & 100k     & 150k       \\
    \multicolumn{1}{l|}{Batch Size}     & 256:0      & 0:720      & 3:48         \\
    \bottomrule
    \end{tabular}
    \caption{\textbf{Hyper-parameters of \thename{}.} The batch size ratio refers to the proportion between multimodal understanding data and visual generation data.}
    \label{tab:Hyper-parameters}
\end{table}

\section{Limitations}
Although our \thename{} achieves promising results with a very small amount of data, the limited data volume still renders our model suboptimal. Furthermore, unlike previous works that leverage large-scale, high-quality datasets such as LAION-aesthetics, we rely solely on the MS-COCO dataset for visual generation. As a result, the quality of generated images, particularly for human faces, remains less refined. Exploring the trade-off between dataset scale and model performance will be a key focus of our future work.

Additionally, while Mamba-2 demonstrates exceptional inference efficiency, its foundational capabilities remain weaker compared to the extensively studied Transformer. Furthermore, Mamba-2 has only been trained on sequences of up to 2048 tokens, limiting its ability to handle ultra-long sequences and hindering its extension to advanced techniques such as Chain-of-Thought (CoT)~\cite{wei2022chain} or reinforcement learning~\cite{guo2025deepseek}. Enhancing Mamba-2's foundational capabilities and its capacity to model ultra-long sequences will be critical areas for future investigation.
\section{Additional Qualitative Results}
\subsection{Multimodal Understanding}
We validate the multimodal understanding capabilities of our \thename{} and other approaches across three aspects: scene description, spatial reasoning, and counting with tricky questions, which are shown in Table 6–8.

\begin{table}[H]
\vspace{-0.3cm}
  \begin{minipage}{0.99\linewidth}
\centering
\scalebox{0.80}{
\begin{tabular}{l p{7.5cm} }
\toprule
 \multicolumn{2}{l}{\bf Spatial Reasoning:}  \\
\midrule
&  \includegraphics[height=3.5cm]{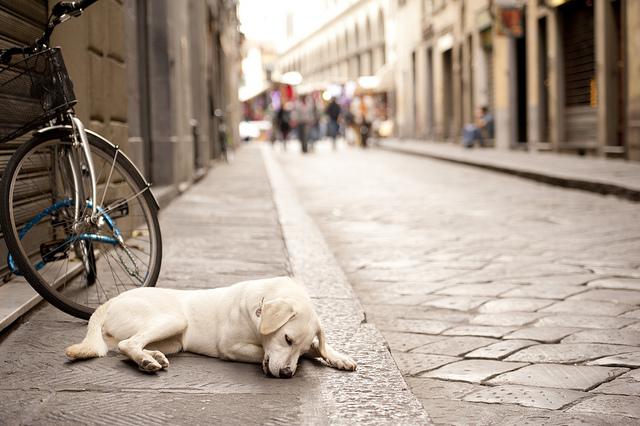} \\
User & Is the bicycle parked to the right of the dog in this image? \\
\midrule
\thename{} & The bicycle is actually parked to the left of the dog, not to the right. \\ \midrule
Show-o & \textcolor{red}{Yes}, the bicycle is parked to the \textcolor{red}{left} of the dog. \\ \midrule
JanusFlow & No. \\
\bottomrule
\end{tabular}
}
\captionof{table}{\thename{} can accurately determine spatial positions, JanusFlow made the correct judgment, but its response was overly concise, while Show-o lacks spatial reasoning capabilities, as indicated by the \textcolor{red}{red} text.}
\label{tab:mmu2}  
  \end{minipage}
\end{table}

\begin{table}[H]
  \begin{minipage}{0.99\linewidth}
\centering
\scalebox{0.80}{
\begin{tabular}{l p{7.5cm} }
\toprule
 \multicolumn{2}{l}{\bf Counting with Tricky Questions:}  \\
\midrule
&  \includegraphics[height=3.5cm]{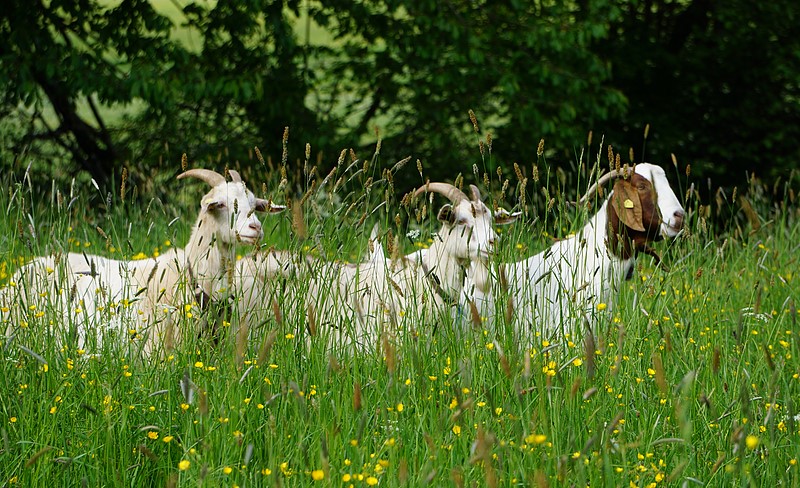} \\
User & The animals in the picture look like cows. How many are there? \\
\midrule
\thename{} & There are three goats in the image, not cows. \\ \midrule
Show-o & There are three \textcolor{red}{cows} in the image. \\ \midrule
JanusFlow & There are three animals in the picture. \\
\bottomrule
\end{tabular}
}
\captionof{table}{\thename{} correctly outputs the quantity while avoiding the user's trap question, whereas Show-o, despite counting correctly, was misled by the question, as indicated by the \textcolor{red}{red} text. On the other hand, JanusFlow's response was not accurate enough.}
\label{tab:mmu3}  
  \end{minipage}
\end{table}

\begin{table}
  \begin{minipage}{0.99\linewidth}
\centering
\scalebox{0.80}{
\begin{tabular}{l p{7.5cm} }
\toprule
 \multicolumn{2}{l}{\bf Scene Description:}  \\
\midrule
&  \includegraphics[height=3.5cm]{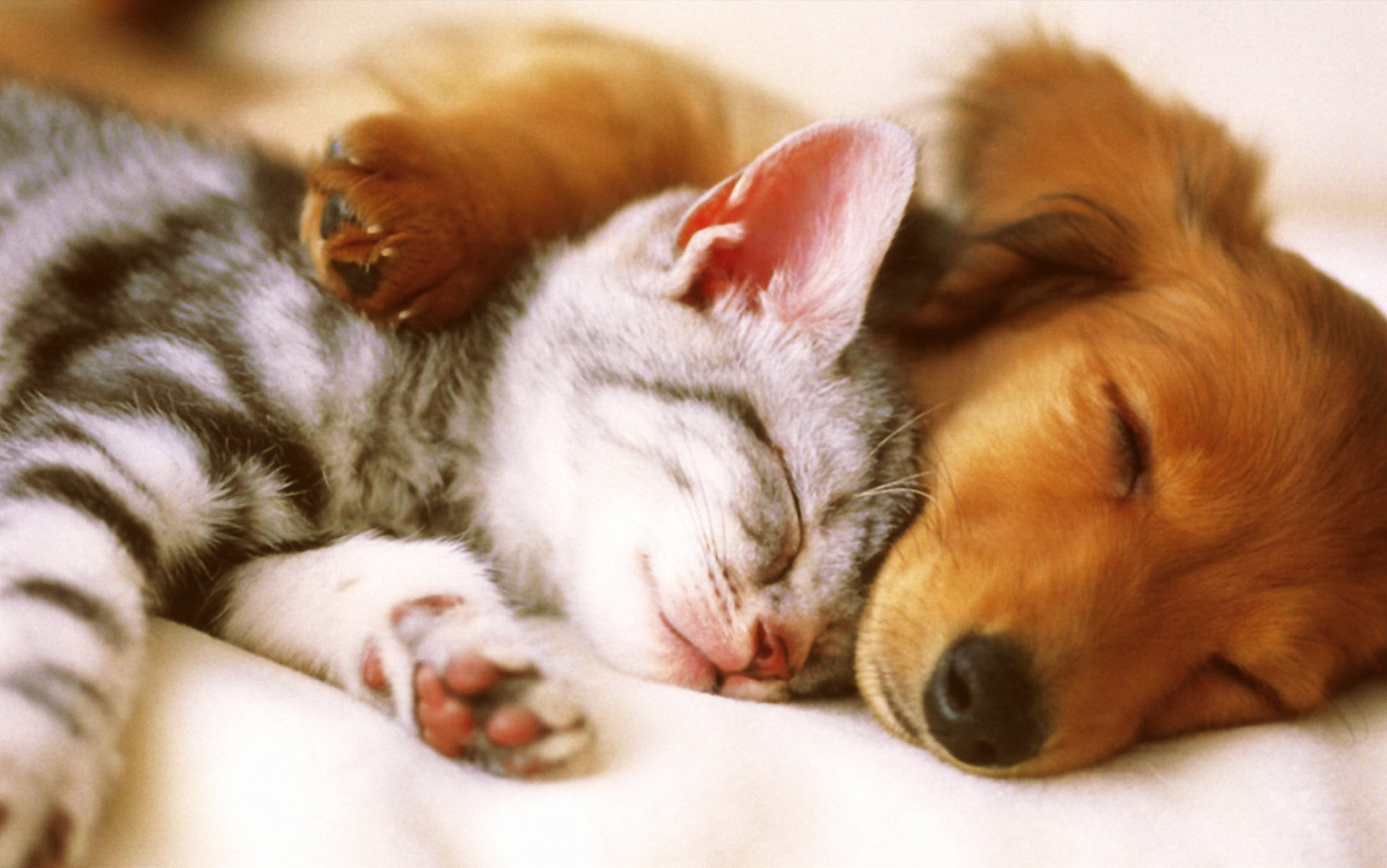} \\
User & Please describe it in detail. \\
\midrule
\thename{} & The image depicts a serene scene of a cat and a dog sharing a peaceful moment together. The cat, exhibiting a gray and white coat, is curled up comfortably, its head resting on the dog's body. The dog, with its brown fur, is curled up in a similar position, with its head resting on the cat's back. Their eyes are gently closed, signifying a deep sleep. The background is a blurred white, allowing the focus to remain on the sleeping feline and its canine friend. The image exudes warmth and affection, showcasing the close bond between the two animals. \\ \midrule
Show-o & The image features a white and \textcolor{red}{brown} cat lying on a bed, curled up and sleeping. The cat is resting its head on a \textcolor{red}{stuffed animal, which is a teddy bear. The teddy bear is positioned on the right side of the cat, with its head resting on the cat's back. The scene captures a peaceful and cozy moment between the cat and the stuffed animal}. \\ 
\bottomrule
\end{tabular}
}
\captionof{table}{\thename{} can accurately describe the information in the scene, whereas Show-o made a mistake about the color of the cat and misidentified the dog as a teddy bear, as indicated by the \textcolor{red}{red} text.}
\label{tab:mmu1}  
  \end{minipage}
  \vspace{-0.3cm}
\end{table}

\subsection{Visual Generation}
\begin{figure*}[h]
    \centering
    \includegraphics[width=0.98\linewidth]{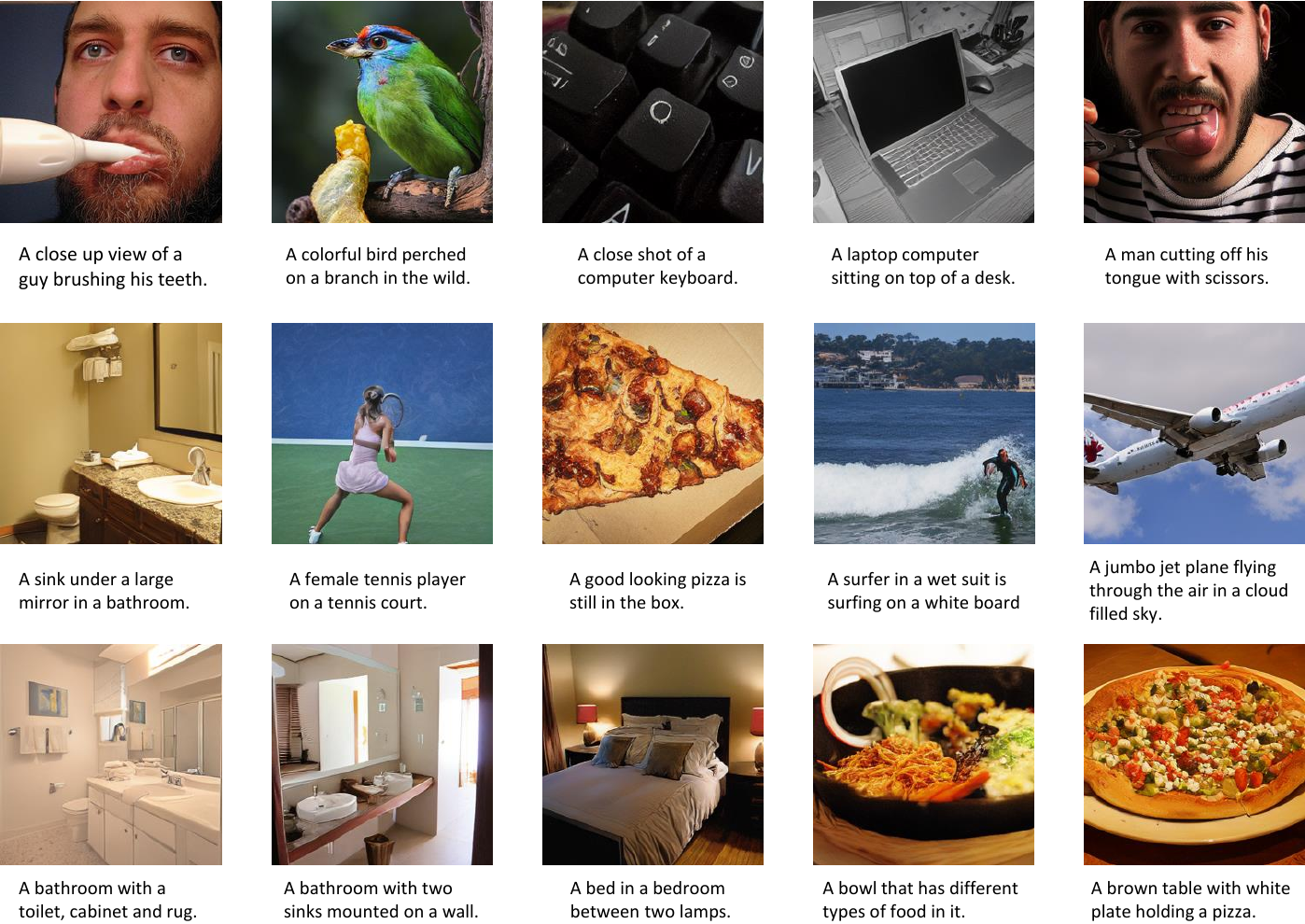}
    \caption{Qualitative results of \thename{} visual generation. Prompts are randomly drawn from the MS-COCO validation set.}
    \label{fig:t2i_qualitative}
\end{figure*}

We provide additional visualization results of visual generation to further validate our generation capabilities, as shown in Fig~\ref{fig:t2i_qualitative}.

\end{document}